\documentclass[letterpaper, 10 pt, conference]{ieeeconf}  

\IEEEoverridecommandlockouts                              

\usepackage{amsmath,amsfonts}
\usepackage{algorithmic}
\usepackage{algorithm}
\usepackage{array}
\usepackage[caption=false,font=normalsize,labelfont=sf,textfont=sf]{subfig}
\usepackage{textcomp}
\usepackage{stfloats}
\usepackage{url}
\usepackage{verbatim}
\usepackage{graphicx}
\usepackage{cite}
\usepackage{tabularx}
\usepackage{multirow}
\usepackage{booktabs}

\title{
LLM-HBT: Dynamic Behavior Tree Construction for Adaptive Coordination in Heterogeneous Robots
}
\author{Chao-ran Wang$^{1}$, Jingyuan Sun$^{*}$, Yan-hui Zhang, Mingyu Zhang, Chang-ju Wu$^{*}$
\thanks{Corresponding author: Chang-ju Wu is with the School of Aeronautic and Astronautics, Zhejiang University, Hangzhou 310027, China. Jingyuan Sun is with Shanghai Huawei Technologies Co., Ltd., Shanghai 201799, China (e-mail: sunjingyuan1@huawei.com,}
}

\begin{document}

\maketitle
\thispagestyle{empty}
\pagestyle{empty}



\maketitle

\begin{abstract}

We introduce a novel framework for automatic behavior tree (BT) construction in heterogeneous multi-robot systems, designed to address the challenges of adaptability and robustness in dynamic environments. Traditional robots are limited by fixed functional attributes and cannot efficiently reconfigure their strategies in response to task failures or environmental changes. To overcome this limitation, we leverage large language models (LLMs) to generate and extend BTs dynamically, combining the reasoning and generalization power of LLMs with the modularity and recovery capability of BTs. The proposed framework consists of four interconnected modules—task initialization, task assignment, BT update, and failure node detection—which operate in a closed loop. Robots tick their BTs during execution, and upon encountering a failure node, they can either extend the tree locally or invoke a centralized virtual coordinator (Alex) to reassign subtasks and synchronize BTs across peers. This design enables long-term cooperative execution in heterogeneous teams. We validate the framework on 60 tasks across three simulated scenarios and in a real-world café environment with a robotic arm and a wheeled-legged robot. Results show that our method consistently outperforms baseline approaches in task success rate, robustness, and scalability, demonstrating its effectiveness for multi-robot collaboration in complex scenarios.

\end{abstract}



\section{Introduction}
Multi-robot systems have demonstrated significant potential in improving operational efficiency by distributing tasks across heterogeneous robots. 
However, existing learning-based methods often struggle to generalize to new environments, as they rely heavily on predefined priors and limited training data~\cite{cai2024transformer}. This limitation becomes critical in dynamic and unstructured settings~\cite{dai2025heterogeneous}, where robots must adapt to failures, environmental changes, and unforeseen conditions. Traditional decision-making frameworks are either too rigid to support online reconfiguration or too brittle to ensure long-term robustness~\cite{lee2025maintaining,zhou2023robust}. Addressing these challenges requires a methodology that enables heterogeneous robots to dynamically reconfigure their task strategies while maintaining scalable and reliable execution.

\begin{figure}[t]
    \centering
    \includegraphics[width=0.48\textwidth]{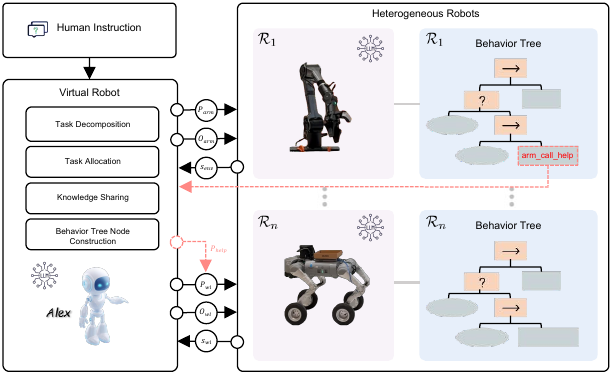}
    \caption{The framework consists of three main components: (i) Human instruction, where natural language commands are provided; (ii) Virtual robot (\textit{Alex}), which decomposes tasks, allocates subtasks, and shares knowledge; and (iii) Heterogeneous robots, including drones, robotic arms, and quadrupeds, which execute their own behavior trees collaboratively.}
    \label{fig1}
\end{figure}

Despite these advances, existing approaches still face fundamental limitations when applied to long-term coordination in heterogeneous multi-robot systems\cite{dai2025heterogeneous}. LLM-based methods, while demonstrating strong reasoning capabilities, often generate task plans in a single-shot manner and lack mechanisms for online correction once execution begins\cite{song2023llm}. This makes them fragile in dynamic or partially observable environments, where unexpected failures or environmental changes are inevitable. Conversely, behavior-tree-based methods offer modularity and recovery mechanisms through structured execution policies\cite{colledanchise2016behavior}, but they rely heavily on manually designed action nodes and predefined task structures\cite{chen2023hybrid}. Such reliance restricts their scalability to unseen tasks and their adaptability to heterogeneous robots with diverse capabilities. More critically, current research has not fully addressed how heterogeneous robots can collaboratively reconstruct and share behavior tree (BT) at runtime to ensure persistent task execution\cite{lee2025maintaining}. These gaps highlight the need for a unified framework that combines the semantic reasoning strength of LLMs with the structural robustness of BT to enable dynamic, distributed, and long-term multi-robot coordination\cite{cai2024transformer}.

To bridge these limitations, we propose a novel framework that integrates large language models with BT to enable dynamic task planning and long-term coordination in heterogeneous multi-robot systems\cite{song2023llm}. At the core of our approach is a centralized allocator that monitors execution progress across robots and detects failures during runtime. When a behavior tree condition node fails, the allocator leverages LLM-based reasoning to infer the most appropriate robot and corresponding actions for recovery\cite{iovino2022survey}. The selected robot then extends its behavior tree by inserting a new subtree that encodes the updated task logic\cite{colledanchise2016behavior}. If the failure requires collaboration, the extended subtree is propagated to additional robots, ensuring distributed consistency while preserving the modularity of BT\cite{chen2023hybrid}. Through this closed-loop cycle of failure detection, reasoning, and tree adaptation, our framework allows heterogeneous robots to continuously refine their execution strategies, complement one another’s functional limitations, and maintain robustness in dynamic and uncertain environments\cite{lee2025maintaining}.

To validate the proposed framework, we design a new benchmark that incorporates diverse simulated scenarios and heterogeneous robot teams, including drones, quadruped robot and robotic arms, tasked with solving complex, long-horizon objectives. Experimental results demonstrate that our method significantly improves task success rates and adaptability compared to existing approaches. 
The main contributions of this work are summarized as follows:

\begin{itemize}
\item 
A dynamic framework that integrates large language model (LLM) reasoning with behavior trees (BTs) for heterogeneous multi-robot coordination. 
\item 
A hybrid centralized–distributed mechanism that enables runtime adaptation through local BT extension and centralized task reallocation.
\item 
A new benchmark covering diverse simulated tasks and a real-world environment, demonstrating the robustness and scalability of the proposed approach. 
\end{itemize}

\section{Related works}

\subsection{Behavior Tree Automatic Design}

Behavior Trees (BTs) have been widely recognized for their modularity, reusability, and interpretability in robotic decision-making~\cite{ghzouli2023behavior}. Recent works have begun to explore automatic BT generation. For instance,~\cite{zhou2024llm} proposed an LLM-based approach for robotic arm control, enabling dynamic addition of operations to handle environmental changes. Similarly,~\cite{izzo2024btgenbot} developed a fine-tuned Llama2 model to generate standard C++ BT strategies. Other studies have integrated optimization-based techniques, such as hierarchical auction algorithms, into BT construction to improve multi-agent task allocation~\cite{heppner2024behavior, tadewos2023automatic}. However, these approaches typically require manual cost function design or operate under simplified assumptions, limiting their adaptability in unstructured environments. In contrast, our method eliminates the need for handcrafted cost functions by leveraging LLM reasoning to interpret environmental observations and dynamically extend BT structures across heterogeneous robots.

\subsection{Multi-robot Planning with LLM}

The integration of large language models (LLMs) into collaborative robotics has recently gained increasing attention. \cite{gong2023lemma} introduced a language-based task allocation method for multi-arm manipulation, while \cite{mandi2024roco} leveraged pre-trained LLMs for high-level communication and low-level path planning. To improve efficiency in complex scenarios, \cite{liu2024coherent} proposed a ``prompt-failure-feedback'' mechanism that enhances task reasoning through historical context. Although effective, such methods may suffer from ambiguous interpretations when facing dynamic environmental changes. Parallel efforts have explored multi-round dialogue mechanisms for task planning \cite{du2023improving, long2024discuss, gu2023maniskill2}, but these studies mainly target homogeneous robot systems. Zhao et al.~\cite{shi2025monte} combined LLMs with Monte Carlo tree search (MCTS) to improve planning efficiency by generating world models as priors. While these methods demonstrate the promise of LLMs for multi-robot collaboration, they lack a formalized and adaptable execution framework. More importantly, the coordination of heterogeneous robots—each with distinct capabilities—remains underexplored.

In summary, prior research on automatic BT design has advanced modular decision-making but often relies on handcrafted cost functions or limited adaptation mechanisms. Meanwhile, recent LLM-based multi-robot planning methods primarily focus on homogeneous systems and lack a structured execution framework. To the best of our knowledge, no existing work has integrated LLM reasoning with dynamic BT construction for heterogeneous multi-robot systems. This study fills that gap by proposing a unified framework that combines the reasoning power of LLMs with the modular adaptability of BTs. Our approach enables online task reconfiguration, distributed execution, and long-term robustness in complex environments.

\section{Preliminaries}

In this section, we formalize the core concepts and notations employed in this work, covering the representation of heterogeneous robot teams, the behavior-tree (BT) formalism, and the specification of action preconditions and effects. These definitions establish the theoretical basis for the LLM-HBT framework developed in the subsequent sections.

\subsection{Heterogeneous Multi-Robot Systems}

We consider a heterogeneous multi-robot system (HMRS) 
\(\mathcal{R} = \{r_1, \dots, r_N\}\), where \(r_i\) is the \(i\)-th robot and \(N\) is the total number of robots. 
Each robot has an action space
\begin{equation}
\mathcal{A}_i = \{a_i^1, \dots, a_i^{m_i}\},
\end{equation}
where \(m_i\) is the number of primitive actions \(a_i^j\) available to \(r_i\) (e.g., \textit{MoveTo}, \textit{Grab}, \textit{TakeOff}). 
Heterogeneity arises from \(\mathcal{A}_i \neq \mathcal{A}_j\) for \(i \neq j\), reflecting differences in morphology and capabilities.  

A task \(\tau\) is represented by a set or sequence of required actions \(\mathcal{A}_\tau \subseteq \bigcup_i \mathcal{A}_i\). 
Assignment of \(\tau\) to \(r_i\) is feasible if \(\mathcal{A}_\tau \subseteq \mathcal{A}_i\). 
For complex tasks, multiple robots may cooperate by combining complementary actions from their respective action spaces.

\subsection{Behavior Trees}

A Behavior Tree (BT) is a rooted directed tree \(T = (V, E, r)\), where \(V\) is the set of nodes, \(E\) defines parent--child edges, and \(r\) is the root. 
Each node \(v \in V\) encodes control logic or an action and returns \textit{Success}, \textit{Failure}, or \textit{Running}. 
Execution proceeds in discrete \textit{ticks} from root to leaves.  

Nodes are:  
\begin{itemize}
    \item \textbf{Control nodes} (e.g., \textit{Sequence}, \textit{Fallback}, \textit{Parallel}) govern flow; \textit{Sequence} fails on the first failing child, \textit{Fallback} succeeds on the first succeeding child.
    \item \textbf{Leaf nodes} represent actions or conditions, interfacing with perception and actuation.
\end{itemize}

A \textit{Sequence} node with children \(\{c_1,\dots,c_n\}\) executes as:
\begin{equation}
\label{eq:seq}
\text{Seq}(c_1,\dots,c_n) =
\begin{cases}
\text{Failure}, & \exists i: c_i = \textit{Failure},\\
\text{Running}, & \exists i: c_i = \textit{Running},\\
\text{Success}, & \forall i: c_i = \textit{Success}.
\end{cases}
\end{equation}
Fallback and Parallel follow analogous semantics.  

Each action node \(a\) is associated with a set of prerequisite conditions, denoted \(\text{Pre}(a)=\{c^{\mathrm{pre}}_1,\dots,c^{\mathrm{pre}}_m\}\), which must hold before execution, and a set of resulting conditions, denoted \(\text{Post}(a)=\{c^{\mathrm{post}}_1,\dots,c^{\mathrm{post}}_m\}\), which are enforced upon successful execution. 
These conditions link BT execution with symbolic reasoning, allowing task states to be explicitly updated and enabling formal analysis of behavior correctness.

\section{Method}
\subsection{Framework Overview}
\begin{figure*}
    \centering
    \includegraphics[width=1\textwidth]{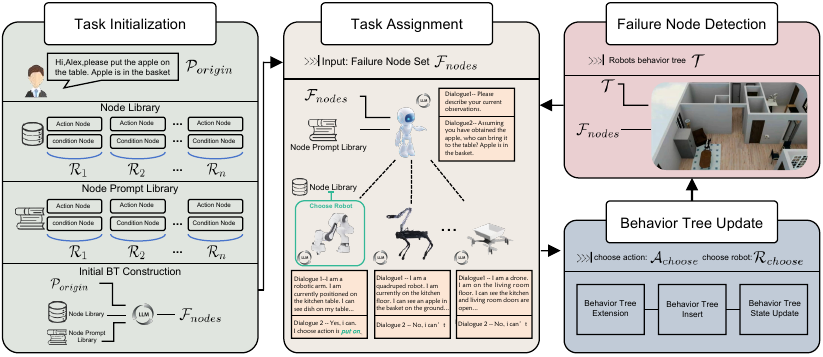}
        \caption{Overview of the proposed framework for automatic behavior tree (BT) construction and adaptation in heterogeneous multi-robot systems. The framework consists of four interconnected modules: (1) \textbf{Task Initialization}, where human instructions are translated into an initial BTs using an LLM and predefined node templates; (2) \textbf{Task Assignment}, where failure nodes trigger the central allocator (Alex) to reassign tasks based on robot capabilities; (3) \textbf{Behavior Tree Update}, where new subtrees are inserted or synchronized across robots to extend task execution strategies; and (4) \textbf{Failure Node Detection}, where robots continuously tick their BTs to monitor execution and identify bottlenecks. These modules operate in a closed-loop cycle, enabling long-term adaptability and robust collaboration in dynamic environments.}
    \label{fig:framework}
\end{figure*}


The framework(Fig.~\ref{fig:framework}) enables automatic BT construction and adaptation in heterogeneous multi-robot systems. Human instructions are first translated into an initial BT using an LLM and a library of node templates (\textbf{Task Initialization}). Robots tick their BTs to monitor progress, reporting unsatisfied conditions as failure nodes (\textbf{Failure Node Detection}). A centralized allocator, \textit{Alex}, collects observations and queries the LLM to assign suitable robots and actions to resolve failures (\textbf{Task Assignment}). The selected actions are integrated into BTs either locally or as synchronized subtrees for collaboration (\textbf{Behavior Tree Update}). Iteratively cycling through these modules allows robots to adapt BTs dynamically to environmental feedback, enabling robust and scalable task execution.

\subsection{Task Initialization}
Before execution, each robot must construct an initial BT based on the human-provided task description. The execution policy of robot $\mathcal{R}_{i}$ is represented by a BT $\mathcal{T}_{i}$, which is updated over time during task execution. Each BT is executed using the $Tick(\cdot)$ function, which propagates control from the root node and returns one of three statuses: \textit{Success}, \textit{Failure}, or \textit{Running}.  

To initialize the BTs, we first define a library of action nodes $\mathcal{A}_{i}$ for each robot $\mathcal{R}_{i}$, along with their associated preconditions and postconditions. A precondition specifies the environmental requirements that must hold before an action can be executed, while a postcondition describes the state transition resulting from the action. For example, the action \textit{grasp(object)} executed by a robotic arm requires the precondition that the object is within reachable distance, and its postcondition is that the object is held by the gripper.

Given a high-level natural language instruction $\mathcal{P}_{origin}$ from a human researcher, we employ an LLM to decompose the task into condition nodes $\mathcal{C}$ that represent the required states of objects and environments. These condition nodes are assembled into an initial BT $\mathcal{T}_{0}$ using predefined templates (e.g., \textit{Sequence} nodes to enforce ordering constraints). The resulting tree is assigned a unique identifier (initialized as $id=1$) and its unsatisfied condition nodes are inserted into the failure node set $\mathcal{F}_{nodes}$. This initialization provides a structured representation of the task, serving as the foundation for subsequent execution and online adaptation.

\begin{algorithm}[H]
\caption{Automatic Design of Behavior Trees for Heterogeneous Multirobots}\label{alg:alg1}
\begin{algorithmic}[1]
\REQUIRE Researcher Prompt $\mathcal{P}_{origin}$
\ENSURE Multirobot behavior tree $\mathcal{T}=\left \{ \mathcal{T}_{1},\cdots,\mathcal{T}_{k}  \right \} $
\STATE Constructing Pre-defined Behavior Tree Nodes $\mathcal{N}$
\STATE Text Generation: $f_{nodes}=llm(\mathcal{P}_{origin},\mathcal{N}\{condition\})$
\STATE Initialize Failure nodes set: $\mathcal{F}_{nodes}.add(f_{nodes})$
\WHILE{$\mathcal{F}_{nodes}\ne \emptyset$}
\STATE $f_{exe}=\mathcal{F}_{nodes}[0]$
\STATE $\mathcal{P}_{Alex}=Q(f_{node},\mathcal{N}\{query\},\mathcal{O})$
\STATE $\mathcal{R}_{choose},\mathcal{P}_{robot}=llm(\mathcal{P}_{Alex})$
\STATE $\mathcal{A}_{choose}=llm(\mathcal{P}_{robot},\mathcal{N}\{\mathcal{R}_{choose}\}.Con)$
\IF{$\mathcal{A}_{choose}\ne None$}
\STATE $\mathcal{T}_{ext}=BT\_Extention(\mathcal{A}_{choose},\mathcal{N}\{\mathcal{R}_{choose}\})$
\STATE $\mathcal{T}_{\mathcal{R}_{choose}}=BT\_Melt(\mathcal{T}_{\mathcal{R}_{choose}},BT_{ext})$
\ELSE{
\STATE \textbf{Warning: This Task Cannot be Completed.}
}
\ENDIF
\FOR{$i\to N$}
\STATE $f_{i},\mathcal{S}_{i}$=\textit{Tick}($\mathcal{T}_{i}$)
\IF{$f_{i} \ne None $}
\STATE $\mathcal{F}_{nodes}.add(f_{i})$
\ENDIF
\IF{$i\equiv \mathcal{R}.index\{\mathcal{R}_{choose}\}$}
\IF{$Tick(f_{exe})\equiv success$}
\STATE $\mathcal{F}_{nodes}.delete(f_{exe})$
\ENDIF
\ENDIF
\ENDFOR
\ENDWHILE
\end{algorithmic}
\label{alg1}
\end{algorithm}

\begin{figure}[t]
   \centering
   \includegraphics[width=0.48\textwidth]{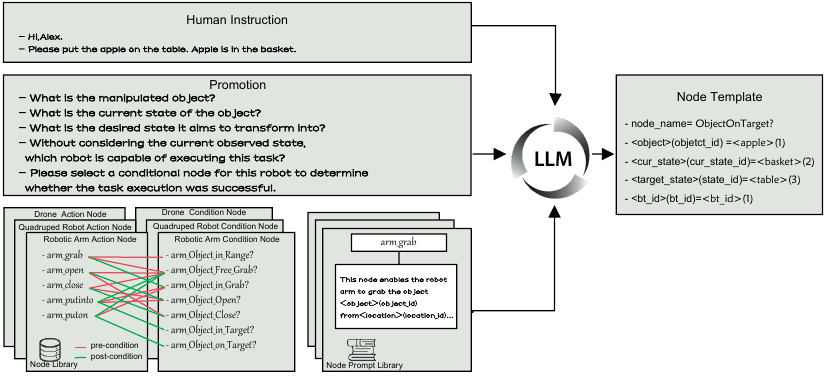}
   \caption{Illustration of the prompt-based behavior tree construction pipeline. Human natural language instructions are first converted into condition/action node prompts, which are then assembled into a behavior tree for robotic execution.}
   \label{fig3}
\end{figure}

\subsection{Task Assignment}
After initialization, the system enters an iterative execution loop in which each robot evaluates its behavior tree using $Tick(\mathcal{T}_{i})$. This process yields both the execution status $\mathcal{S}_{i}$ and, when applicable, a failed condition node $f_{i}$. Not all failures require external intervention: a robot first attempts to resolve its own failure by extending its BT using its internal action set $\mathcal{A}_{i}$. When the robot determines that its capabilities are insufficient to satisfy the failed condition, it triggers a dedicated \textit{call\_help} action node.  

To handle such requests, we introduce a virtual centralized allocator, denoted as \textit{Alex}. The role of \textit{Alex} is to mediate information sharing among robots and to allocate tasks dynamically when local resolution is not feasible. Each call\_help query is formalized as a prompt containing three elements: (1) the failed condition node $f_{i}$, (2) the preconditions and postconditions associated with the node, and (3) the robot’s partially observable state $\mathcal{O}_{i}$, which includes both environmental observations and relational information between objects.  

Given this structured prompt, the LLM reasons about which robot $\mathcal{R}_{j}$ is best suited to resolve the failure, considering the action sets $\mathcal{A}_{j}$ available across the team. \textit{Alex} then returns a task assignment $\mathcal{P}_{robot}$ and the corresponding candidate action node $\mathcal{A}_{choose}$. The selected robot incorporates this node into its own BT by extending it with a new subtree, while the requesting robot updates its BT to synchronize with the assignment.  

Through this mechanism, task allocation is achieved in a distributed manner: robots autonomously manage local failures when possible, and rely on \textit{Alex} only when inter-robot collaboration is required. This hybrid allocation strategy ensures scalability and reduces unnecessary communication overhead while still enabling efficient cooperation in heterogeneous multi-robot teams.

\begin{figure}
    \centering
    \includegraphics[width=0.48\textwidth]{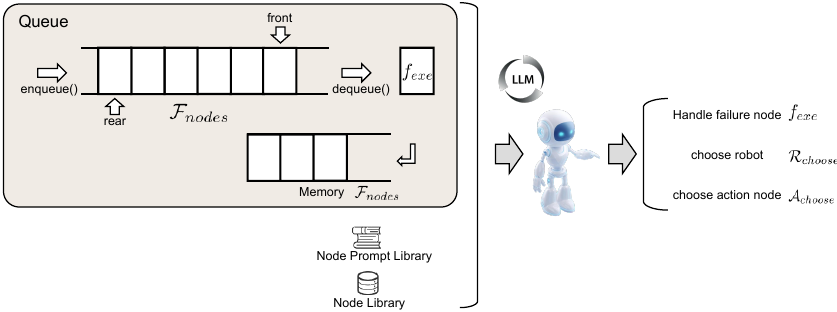}
    \caption{Process of failure node detection, storage, and reuse in behavior tree execution.
}
    \label{fig4}
\end{figure}

\subsection{Behavior Tree Update}

We define both \textit{a priori} and \textit{a posteriori} conditions for each robot action. The \textit{a priori} condition specifies the environmental requirements that must be satisfied before an action can be executed (e.g., an object must be within the graspable range of a robotic arm before a grasp action is attempted). The \textit{a posteriori} condition describes the resulting state transition after the action is completed (e.g., the object is now held by the gripper). 

The update of the BT structure is primarily realized through two operations: \textit{extension} and \textit{insertion}. Let $\mathcal{A}_{i}$ denote the set of executable action nodes of robot $\mathcal{R}_{i}$, and $\mathcal{C}_{i}$ denote its condition nodes. During task execution, each robot executes its BT $\mathcal{T}_{i}$ through $Tick(\mathcal{T}_{i})$, which returns an execution status $\mathcal{S}_{i}$ and, if applicable, a failure node $f_{i}$. When a failure node is encountered, the system constructs a candidate extended subtree $\mathcal{T}_{ext}$, which is then inserted into $\mathcal{T}_{i}$ in place of $f_{i}$ (Algorithm~\ref{alg1}, Line 10-11). The purpose of this process is to transform the outcome of the failed condition into a successful state by either exploiting the robot’s intrinsic capabilities or by delegating the task to other robots. 

In our framework, delegation is managed by the centralized virtual allocator \textit{Alex}, which maintains a shared view of the robots and their environmental states. When a failure node $f_{i}$ is reported, \textit{Alex} collects contextual information including the label, object, location, and target state of the failed condition, along with the reporting robot’s partially observable state. Based on this information, \textit{Alex} identifies whether the failure can be addressed locally by robot $\mathcal{R}_{i}$ or should be resolved by another robot $\mathcal{R}_{j}$ with the appropriate a posteriori effect. If the latter holds, \textit{Alex} instructs $\mathcal{R}_{j}$ to construct an extended subtree $\mathcal{T}_{ext}^{j}$, which is integrated into $\mathcal{T}_{j}$ with high execution priority to ensure timely recovery. 

\begin{figure}
    \centering
    \includegraphics[width=0.48\textwidth]{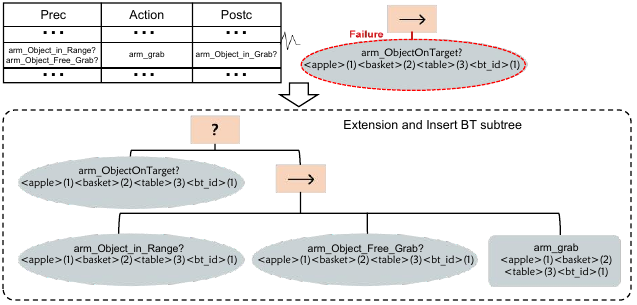}
    \caption{Behavior tree extension and insertion process. 
    When a failure node is encountered, the system generates an extended subtree through LLM-based reasoning. This subtree is then inserted into the existing behavior tree, enabling the robot to recover from failures and continue task execution.}
    \label{fig:BT_extension}
\end{figure}

Two collaborative update cases arise from this mechanism. (1) \textbf{Independent extension}: If the robot can resolve the failed condition using its own action set, the subtree $\mathcal{T}_{ext}$ is directly merged into $\mathcal{T}_{i}$ by replacing $f_{i}$ (Fig.~\ref{fig:BT_extension}a). This allows the robot to recover autonomously from execution failures. (2) \textbf{Delegated extension}: If the failure node cannot be resolved locally, \textit{Alex} assigns the task to a peer robot $\mathcal{R}_{j}$. In this case, $\mathcal{R}_{i}$ temporarily suspends execution at $f_{i}$ while monitoring its status, whereas $\mathcal{R}_{j}$ incorporates the delegated subtree $\mathcal{T}_{ext}^{j}$ at the root of its behavior.

\subsection{Failure Node Detection}
During task execution, each robot $\mathcal{R}_{i}$ continuously evaluates its behavior tree $\mathcal{T}_{i}$ through the standard $Tick(\cdot)$ operation. A \textit{Failure} status indicates that the corresponding condition node cannot be satisfied under the current environmental state. The failed node is then recorded as a failure node $f_{i}$ and inserted into a dedicated failure queue $\mathcal{F}_{nodes}$ for further processing (Algorithm~\ref{alg:alg1}, Lines 15-25).  

Unlike conventional BT implementations where failures simply propagate upward and potentially terminate execution, our framework explicitly monitors and stores these nodes. This mechanism provides two key advantages: (1) it allows systematic identification of execution bottlenecks that require intervention, and (2) it serves as the trigger for invoking the extension process to recover from the failure.  

Once a failure node is detected, the information is forwarded to the centralized allocator \textit{Alex}, which maintains a shared view of the environment and robot states. By leveraging this global knowledge, \textit{Alex} determines whether the failure can be resolved by the same robot through a local action extension, or whether the task requires delegation to another robot with the appropriate capabilities. In the latter case, \textit{Alex} coordinates the integration of an extended subtree into the corresponding robots’ behavior trees, ensuring that the overall mission continues to progress.  

Through this mechanism, failure detection does not represent the termination of execution but instead acts as an adaptive decision point that drives the dynamic reconstruction of behavior trees. By systematically recording and processing failure nodes, the system achieves resilience against unexpected disturbances and maintains long-term task execution in dynamic environments.

\section{Experiments}
\subsection{Simulation Experiments}
We first evaluated the proposed framework in simulation using the Behavior-1K dataset, which provides diverse task descriptions spanning navigation, object manipulation, and cooperative missions. To construct a representative yet tractable benchmark, we sampled 20 tasks per group and report their detailed execution results in Table \ref{table:simulation_20tasks_horizontal}. Each entry in the table corresponds to the number of behavior-tree ticks (execution steps) required to complete the task, while “x” denotes a failure case. The sampled tasks cover varying action horizons, ranging from 2 to 20 steps across different scenarios. To assess scalability, we designed three scenarios with increasing heterogeneity: (i) a single quadruped robot, (ii) a quadruped robot paired with a drone, and (iii) a quadruped robot, a drone, and a robotic arm. For each scenario, the 20 tasks were executed over five independent trials with randomized initial positions of both robots and objects, and the averaged results are reported.

We further evaluated the proposed BT-based execution framework on a heterogeneous multi-robot team comprising a robotic arm, a quadruped robot, and a drone. Each robot is modeled with a library of primitive actions, where every action is specified by explicit preconditions and postconditions to maintain logical consistency (see Table~\ref{table:bt_nodes}). The overall number of available actions and conditions for each robot is summarized in Table~\ref{table:bt_summary}. At the beginning of each trial, all robots were initialized in nominal configurations. Actions were executed only when their preconditions were satisfied, and postconditions were continuously monitored to verify success or detect failures. This design ensures that the BT framework strictly enforces logical dependencies across heterogeneous robots while enabling both sequential and coordinated execution.

\begin{table}[t]
\centering
\caption{Behavior Tree (BT) Nodes for Robot Arm, Qudrupted Robot, and Drone}
\label{table:bt_nodes}
\scriptsize
\resizebox{\columnwidth}{!}{%
\begin{tabular}{lll}
\toprule
Precondition & Action & Postcondition \\
\midrule
\multicolumn{3}{l}{\textbf{Robot Arm}} \\
\begin{tabular}[c]{@{}l@{}}ArmObjectFreeGrab?\\ ArmContainClose?\end{tabular} & Open & ArmContainOpen? \\
\begin{tabular}[c]{@{}l@{}}ArmObjectInRange?\\ ArmContainOpen?\end{tabular} & Grab & ArmObjectInGrab? \\
\begin{tabular}[c]{@{}l@{}}ArmObjectFreeGrab?\\ ArmContainOpen?\end{tabular} & Close & ArmContainClose? \\
\begin{tabular}[c]{@{}l@{}}ArmObjectInGrab?\\ ArmContainOpen?\end{tabular} & PutInto & \begin{tabular}[c]{@{}l@{}}ArmObjectInTarget?\\ ArmObjectFreeGrab?\end{tabular}\\
ArmObjectInGrab? & PutOn & \begin{tabular}[c]{@{}l@{}}ArmObjectOnTarget?\\ ArmObjectFreeGrab?\end{tabular}\\
\midrule
\multicolumn{3}{l}{\textbf{Qudrupted Robot}} \\
QuadFreePath? & MoveToNoObject & QuadInRangeNoObject? \\
\begin{tabular}[c]{@{}l@{}}QuadInRangeNoObject?\\ QuadObjectFreeGrab?\end{tabular} & MoveToWithObject & QuadInRangeNoObject? \\
\begin{tabular}[c]{@{}l@{}}QuadObjectFreeGrab?\\ ContainClose?\end{tabular} & Open & QuadContainOpen? \\
\begin{tabular}[c]{@{}l@{}}QuadCanGetObject?\\ QuadInRangeNoObject?\\ QuadObjectFreeGrab?\\ QuadContainOpen?\end{tabular} & Grab & QuadObjectInGrab? \\
\begin{tabular}[c]{@{}l@{}}QuadObjectFreeGrab?\\ QuadContainOpen?\end{tabular} & Close & QuadContainClose? \\
\begin{tabular}[c]{@{}l@{}}QuadInRangeWithObject?\\ QuadContainOpen?\\ QuadObjectInGrab?\end{tabular} & PutInto & \begin{tabular}[c]{@{}l@{}} QuadObjectInTarget?\\ QuadObjectFreeGrab?\end{tabular} \\
\begin{tabular}[c]{@{}l@{}}QuadInRangeWithObject?\\ QuadObjectInGrab?\end{tabular} & PutOn & \begin{tabular}[c]{@{}l@{}} QuadObjectOnTarget?\\  QuadObjectFreeGrab?\end{tabular}\\
\midrule
\multicolumn{3}{l}{\textbf{Drone}} \\
\begin{tabular}[c]{@{}l@{}}DroneObjectInBasket?\\ DroneOnGround?\end{tabular} & TakeOffWithObject & DroneInAirWithObject? \\
\begin{tabular}[c]{@{}l@{}}DroneObjectInBasket?\\ DroneInAirWithObject?\\ DroneInRangeWithObject?\end{tabular} & LandOnWithObject & \begin{tabular}[c]{@{}l@{}}DroneAtTargetWithObject?\\ DroneOnGround?\end{tabular} \\
\begin{tabular}[c]{@{}l@{}}DroneObjectInBasket?\\ DroneInAirWithObject?\\ DronePathFree?\end{tabular} & MoveToWithObject & DroneInRangeWithObject? \\
DroneOnGround? & TakeOffNoObject & DroneInAirNoObject? \\
DroneInRangeNoObject? & LandOnNoObject & \begin{tabular}[c]{@{}l@{}}DroneAtTargetNoObject?\\ DroneOnGround?\end{tabular} \\
\begin{tabular}[c]{@{}l@{}}DronePathFree?\\ DroneInAirNoObject?\\ DronePathFree?\end{tabular} & MoveToNoObject & DroneInRangeNoObject? \\
\bottomrule
\end{tabular}%
}
\end{table}

\begin{table}[t]
\centering
\caption{Summary of action nodes and conditions.}
\label{table:bt_summary}
\footnotesize
\begin{tabular}{lccc}
\toprule
Robot & Actions Number & conditions Number \\
\midrule
Robotic Arm & 5 & 6  \\
Quadruped   & 7 & 8  \\
Drone       & 6 & 7  \\
\bottomrule
\end{tabular}
\end{table}

For a fair comparison, both MCTS and LLM-MCTS were configured with an identical rollout budget of 500 simulations per decision step and a maximum search depth of 20. These hyperparameters were selected to balance computational feasibility with performance and were kept consistent across all scenarios. We benchmark our method (\textbf{LLM-HBT}) against two representative baselines: (i) \textbf{MCTS}, which performs action planning solely through Monte Carlo Tree Search, and (ii) \textbf{LLM-MCTS}, which augments MCTS with LLM-generated world models.

For quantitative evaluation, we adopt two widely used metrics: \textbf{Success Rate (SR)} and \textbf{Average Steps (AS)}. Let $N$ denote the total number of tasks, and let $s_{i} \in \{0,1\}$ indicate whether task $i$ is successfully completed. The success rate is defined as
\begin{equation}
    SR = \frac{1}{N} \sum_{i=1}^{N} s_{i}.
\end{equation}

To measure efficiency, let $k_{i}$ denote the number of behavior-tree ticks required to complete task $i$. The average steps is then defined as
\begin{equation}
    AS = \frac{1}{N} \sum_{i=1}^{N} k_{i}.
\end{equation}

Together, these two metrics characterize both the \emph{effectiveness} of different methods (task completion capability) and their \emph{efficiency} (execution cost in terms of steps).

Table~\ref{tab:overall_results} summarizes the results across all scenarios. Our method consistently achieves a $100\%$ task success rate, while the baselines exhibit substantial degradation as heterogeneity and task complexity increase. In Scenario~1, all methods maintain relatively high success rates (MCTS: $95\%$, LLM-MCTS: $90\%$, LLM-HBT: $100\%$). However, in Scenario~2, both baselines drop to $55\%$, whereas LLM-HBT continues to achieve perfect success. In the most challenging Scenario~3, where three heterogeneous robots must coordinate, the baselines succeed in only $40\%$ of tasks, whereas LLM-HBT still achieves $100\%$. 

It is important to emphasize that the perfect success rate of LLM-HBT does not imply that the tasks are trivial or that the method overfits to the benchmark. Instead, it highlights the intrinsic advantage of behavior trees: once a failure node is detected, the tree can always be extended until a feasible action sequence is identified, either by the current robot or through task reallocation to peers. The primary benefit of LLM-HBT lies in its efficiency, as reflected by the reduced number of average steps compared with baselines. For example, in Scenario~3 our method achieves a lower average step count ($8.4$) compared to MCTS ($8.80$) and LLM-MCTS ($9.00$), demonstrating the effectiveness of LLM reasoning in guiding tree expansion and reducing redundant search. Overall, these results confirm that LLM-HBT provides both robust task completion and improved execution efficiency in heterogeneous multi-robot systems.

\begin{figure*}
    \centering
    \includegraphics[width=1\textwidth]{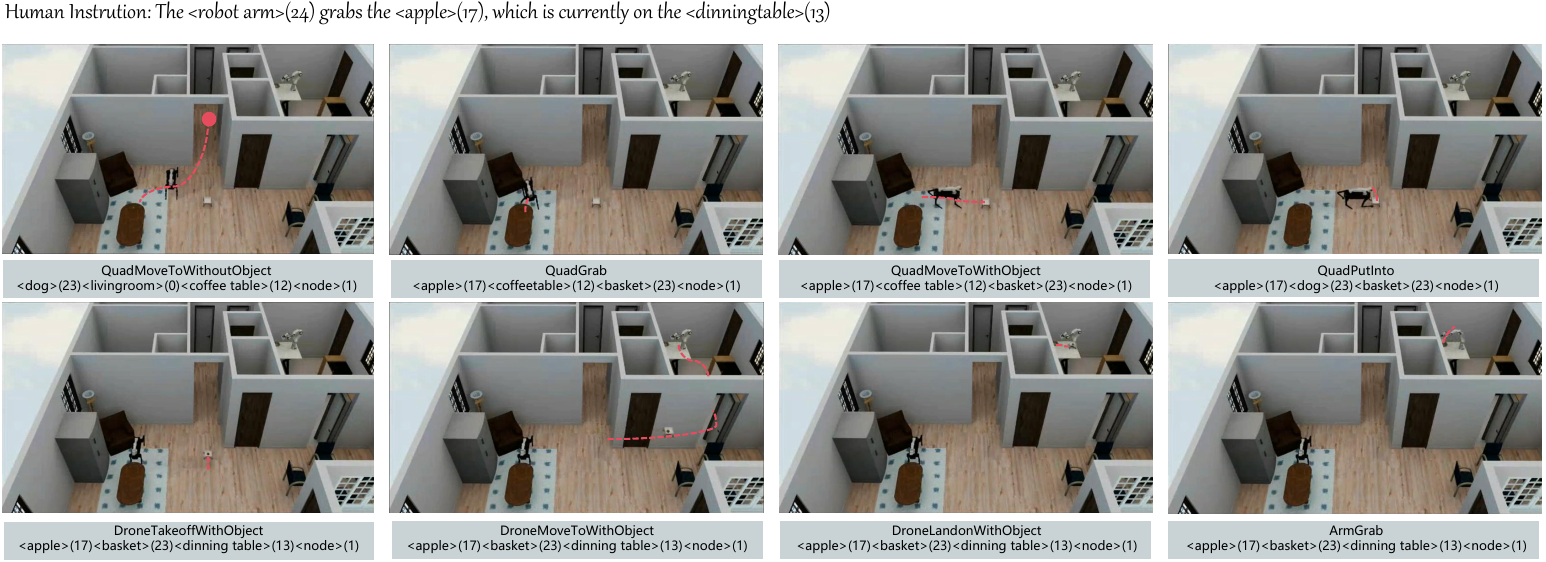}
    \caption{Illustration of Scenario~3 with three robots collaboratively executing a task. The quadruped robot navigates to the table to grab an apple, while the final goal is for the robotic arm to grasp the apple}
    \label{fig1}
\end{figure*}

Table~\ref{table:simulation_20tasks_horizontal} presents the outcomes of 20 tasks per group across three methods: MCTS, LLM-MCTS, and our proposed LLM-HBT. Each cell indicates the performance metric for a specific task, with ``x'' representing task failure. 

Table~\ref{table:simulation_20tasks_horizontal} presents the outcomes of 20 tasks per group across three methods: MCTS, LLM-MCTS, and our proposed LLM-HBT. Each cell indicates the number of execution steps for a specific task, with ``x'' denoting task failure. 

Overall, the proposed LLM-HBT consistently outperforms both baselines across all three groups. In Group~1, LLM-HBT completed all tasks with only minor deviations in execution steps, achieving lower or comparable step counts relative to MCTS and LLM-MCTS. Notably, tasks T12 and T16 that failed under MCTS or LLM-MCTS were successfully executed by LLM-HBT, demonstrating improved robustness. In Group~2, our method successfully completed several tasks that either baseline failed (e.g., T3, T4, and T20), while maintaining efficient execution in tasks completed by all methods. Group~3 further highlights the advantage of LLM-HBT in complex heterogeneous settings: while MCTS and LLM-MCTS failed the majority of tasks (marked as ``x''), our method achieved successful execution in nearly all tasks, often with lower or comparable step counts (e.g., T4, T7, T12, and T20). 

These results confirm that embedding LLM reasoning within the BT framework enables more reliable and adaptive task execution. LLM-HBT not only improves robustness in the face of task failures but also facilitates coordinated, sequential, and parallel execution across heterogeneous robots, even in the most challenging scenarios where baselines frequently fail.

\begin{table*}[t]
\centering
\caption{Simulation results for 20 tasks per group (horizontal: tasks, vertical: methods).}
\label{table:simulation_20tasks_horizontal}
\footnotesize
\renewcommand{\arraystretch}{1.1}
\begin{tabular}{l|cccccccccccccccccccc}
\toprule
\hline
\textbf{Group 1} & T1 & T2 & T3 & T4 & T5 & T6 & T7 & T8 & T9 & T10 & T11 & T12 & T13 & T14 & T15 & T16 & T17 & T18 & T19 & T20 \\
\midrule
MCTs                & 2 & 2 & 2 & 4 & 4 & 4 & 6 & 4 & 4 & 4 & 4 & x & 4 & 6 & 3 & 3 & 6 & 3 & 7 & 3 \\
LLM-MCTs            & 2 & 2 & 2 & 4 & 4 & 4 & 6 & 4 & 4 & 4 & 4 & 7 & 4 & x & 3 & x & 6 & 4 & 7 & 3 \\
Ours       & 2 & 2 & 2 & 3 & 4 & 4 & 6 & 4 & 5 & 5 & 4 & 7 & 4 & 6 & 3 & 1 & 6 & 3 & 7 & 3 \\ \hline
\midrule
\textbf{Group 2} & T1 & T2 & T3 & T4 & T5 & T6 & T7 & T8 & T9 & T10 & T11 & T12 & T13 & T14 & T15 & T16 & T17 & T18 & T19 & T20 \\
\midrule
MCTs           & 2 & 3 & x & x & x & 2 & 4 & x & 7 & 9 & x & x & 4 & 4 & x & x & 5 & 6 & x & 8 \\
LLM-MCTs       & 2 & 3 & x & x & x & 2 & 4 & 6 & 6 & 9 & x & x & 5 & 5 & x & x & 6 & x & x & 9 \\
Ours  & 2 & 3 & 3 & 3 & 4 & 2 & 4 & 4 & 2 & 7 & 7 & 7 & 3 & 2 & 7 & 7 & 3 & 4 & 10 & 8 \\ \hline
\midrule
\textbf{Group 3} & T1 & T2 & T3 & T4 & T5 & T6 & T7 & T8 & T9 & T10 & T11 & T12 & T13 & T14 & T15 & T16 & T17 & T18 & T19 & T20 \\
\midrule
MCTs            & 6 & 3 & 5 & x & x & 2 & x & 6 & 7 & 7 & 5 & x & x & x & x & x & x & x & x & x \\
LLM-MCTs        & 6 & 3 & 5 & x & x & 2 & x & 6 & 7 & 7 & 5 & x & x & x & x & x & x & x & x & x \\
Ours   & 4 & 3 & 7 & 20 & 3 & 2 & 15 & 10 & 7 & 7 & 5 & 19 & 12 & 6 & 11 & 2 & 6 & 5 & 7 & 13 \\ \hline
\bottomrule
\end{tabular}
\end{table*}

\begin{table}
\centering
\caption{Overall performance of different methods across three scenarios (SR: task success rate, Avg. Steps: average number of action steps).}
\label{tab:overall_results}
\begin{tabular}{c|cc|cc|cc}
\hline
\multirow{2}{*}{Method} & \multicolumn{2}{c|}{Scenario1}     & \multicolumn{2}{c|}{Scenario2}     & \multicolumn{2}{c}{Scenario3}      \\ \cline{2-7} 
                        & \multicolumn{1}{c|}{SR(\%)} & AS   & \multicolumn{1}{c|}{SR(\%)} & AS   & \multicolumn{1}{c|}{SR(\%)} & AS   \\ \hline
MCTs                    & 95                          & 3.95 & 55                          & 4.91 & 35                          & 8.80 \\
LLM-MCTs                & 90                          & 4.11 & 55                          & 5.18 & 35                          & 9.00 \\ \hline
LLM-HBT                  & \textbf{100}                & \textbf{4.05} & \textbf{100}       & \textbf{5.05}  & \textbf{100}      & \textbf{8.4} \\ \hline
\end{tabular}
\end{table}

\subsection{Real-World Deployment}
To further verify the stability and effectiveness of our approach in physical environments, we conduct real-world experiments in a café-like setting. The environment includes two heterogeneous robots: (i) a fixed robotic arm mounted on the bar counter and (ii) a wheeled–legged robot equipped with a tray for transporting objects. The task is initialized with a bottle held in a human hand. The objective is for the robotic arm and the wheeled–legged robot to cooperate in placing the bottle on the counter.

\begin{figure}[htbp]
    \centering
    \includegraphics[width=0.5\textwidth]{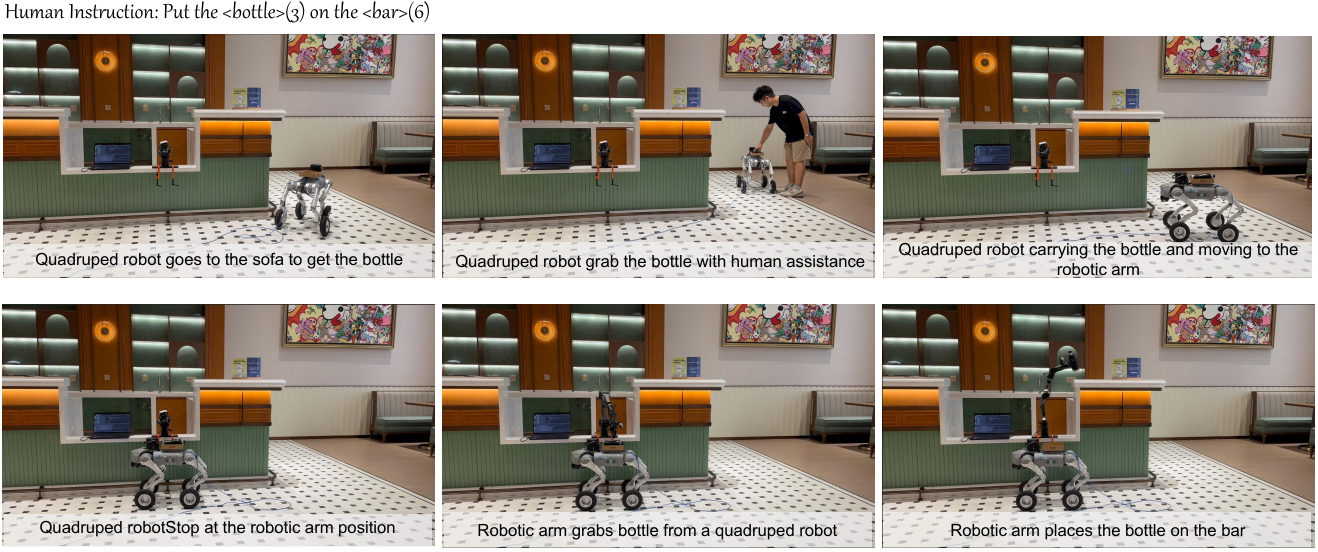}
    \caption{Real-world deployment in a café environment. A wheeled-legged robot equipped with a basket transports a bottle handed over by a human, while a Robotic Arm mounted on the bar counter takes over and places the bottle onto the designated spot.}
    \label{fig:realworld_cafe}
\end{figure}

During execution, the robotic arm first establishes the precondition that the bottle is within its graspable workspace. Since this condition is not initially satisfied, the system extends the behavior tree by allocating the task to the wheeled–legged robot. The robot navigates toward the human, receives the bottle, and delivers it to the counter. The robotic arm then extends its own behavior tree to complete the final grasp-and-place operation. This process demonstrates the ability of the system to dynamically reconfigure behavior trees and allocate subtasks in real time, even under partial observability and physical uncertainties such as motion noise and human variability.

Across ten repeated trials, the proposed method consistently completed the collaborative task without failure, validating that the integration of LLM reasoning and behavior tree expansion enables heterogeneous robots to execute complex, long-horizon tasks robustly in real-world environments. Although LLM-HBT consistently achieved near-perfect success rates, this does not imply that the tasks were trivial. The reason is that behavior trees can be iteratively extended until a feasible solution is found, ensuring eventual task completion. However, this comes at the cost of additional steps or tree updates. In extreme scenarios with severe sensory noise or delayed communication, failure may still occur, which we leave as future work to explore more robust recovery mechanisms.
\section{Conclusion}

We presented \textbf{LLM-HBT}, a novel framework that fuses large language model reasoning with behavior-tree execution to enable long-term, adaptive coordination in heterogeneous multi-robot teams. The system employs a closed-loop pipeline—task initialization, failure detection, LLM-guided task assignment (via a virtual allocator \textit{Alex}), and online BT extension—which allows robots to locally recover or reassign subtasks at runtime. Extensive simulation and a real-world café deployment demonstrate that LLM-HBT substantially improves task success rates and execution efficiency compared to strong baselines, while maintaining modularity and interpretability. Limitations include LLM inference latency and the current scope of real-world validation; future work will focus on latency-aware designs, communication-efficient decentralization, and robustness under perceptual uncertainty. Overall, LLM-HBT offers a practical and scalable approach for adaptive multi-robot coordination at the intersection of language reasoning and embodied control.

\bibliographystyle{IEEEtran}
\bibliography{ref}

\end{document}